\documentclass{ceurart}

\usepackage{tabularx}
\usepackage{subcaption}
\usepackage{float}
\usepackage{enumitem}
\usepackage{multicol}
\usepackage{tikz}
\usetikzlibrary{positioning,arrows.meta,fit,backgrounds,calc}
\usepackage{cleveref}
\usepackage{listings}
\usepackage{xcolor}

\lstdefinestyle{promptstyle}{
  basicstyle=\footnotesize,
  breaklines=true,
  breakatwhitespace=true,
  columns=flexible,
  keepspaces=false,
  showstringspaces=false,
  frame=none,
  xleftmargin=0pt,
  xrightmargin=0pt,
  aboveskip=2pt,
  belowskip=2pt,
  numbers=none
}

\restylefloat{table}
\restylefloat{figure}

\newcommand{\sysname}{GRACE}      % rename here to change the system name everywhere
\newcommand{\keyresult}[1]{{\boldmath\textbf{#1}}}  % bold an important empirical result (text + inline math)

\begin{document}

% ===== Front matter (CEUR-WS) =====
\copyrightyear{2026}
\copyrightclause{Copyright for this paper by its authors.
  Use permitted under Creative Commons License Attribution 4.0
  International (CC BY 4.0).}

\conference{ITADATA2026: The 5$^{\text{th}}$ Italian Conference on Big Data
  and Data Science, November 10--12, 2026, Bari, Italy}

% NOTE: pdfx (loaded by ceurart) re-reads this title from main.xmpdata at class-load
% time, before any macro is defined, so the title must be plain text -- no \sysname,
% no $...$ math. The system name is GRACE; use \sysname{} freely in the body.
\title{GRACE: Parallel Best-of-N Multi-Agent Analysis of Closed-Form Questions over Dataframes with Guardrail Selection and Adaptive Retry}

\author[1]{Fabio Picariello}[email=fabio.picariello@eng.it]
\cormark[1]
\author[1]{Salvatore Cipolla}[email=salvatore.cipolla@eng.it]
\cormark[1]

\author[1]{Susanna Bonura}[email=susanna.bonura@eng.it]
\author[1]{Vito Morreale}[email=vito.morreale@eng.it]

\address[1]{Engineering Ingegneria Informatica S.p.A., Piazzale dell’Agricoltura 24, 00144 Roma, Italy}
\cortext[1]{Corresponding author}

% =====================================================================
\begin{abstract}
LLM agents answer closed-form questions over a tabular dataframe by writing and executing Python, almost always as a single rollout: one interpretation, one script, one result, accepted without comparison. The few pipelines that produce several candidates reconcile them by answer-agreement voting, which needs a single comparable scalar and always returns one, even when every candidate is wrong. We present \sysname{} (Generative, Ranked, Adaptive Code Ensemble), a production-oriented pipeline that instead runs parallel best-of-$N$ selection specialised to dataframe code. $N$ agents run concurrently under a strict, machine-checkable output contract, diversified by persona hints, per-retry rotation, and a two-regime temperature schedule. Rather than vote, \sysname{} judges the executed artefacts: a pairwise tournament pre-filters the pool and a domain-specialised guardrail either selects the best candidate or, when none is defensible, rejects the whole pool and abstains with a diagnosis (an option voting structurally lacks), which in turn drives an adaptive retry with per-agent feedback. On InfiAgent-DABench (DAEval, $M{=}246$), using Qwen3.6-35B-A3B as the backbone, \sysname{} raises answer match from \keyresult{$39\%$} for a single-rollout baseline to \keyresult{$71\%$} (\keyresult{$+32$\,pp}, \keyresult{$1.8\times$}), recovering \keyresult{$53\%$} of previously wrong answers. On the same DAEval task, the smaller dense Qwen3.5-9B lifts answer match from \keyresult{$38\%$} to \keyresult{$63\%$} (\keyresult{$+25$\,pp}, \keyresult{$1.7\times$}). A controlled $3{\times}3$ study that sweeps the number of parallel agents against the retry budget finds, on the 35B backbone, the two levers largely substitutable---once either one is generous, the other adds little---which pins down a cheap operating point for deployment. The same directional pattern replicates on a second, independent benchmark, DataBench, again on the Qwen3.6-35B-A3B backbone, where answer match rises by \keyresult{$15$} percentage points (from \keyresult{$70\%$} to \keyresult{$85\%$}) across $400$ sampled questions.
\end{abstract}


\maketitle

% =====================================================================
\section{Introduction}
\label{sec:introduction}

Natural-language interfaces to tabular data increasingly delegate the analytical step to an LLM code agent: given a dataframe and a question in natural language, the agent writes Python, executes it in a sandbox, and returns what the code produced: a value, a table, or a chart. We target the \emph{closed-form} case: each question has a single well-defined answer computed by executable Python over one in-memory dataframe, the formulation measured by benchmarks such as InfiAgent-DABench~\cite{hu2024infiagent} and instantiated by a wave of analysis agents that \Cref{sec:related} surveys. Two assumptions recur across them. First, each commits to a single solution trajectory: one reading of an often under-specified question, one script, one result, accepted with no comparison against alternatives; even the systems that iterate refine that single trajectory rather than choosing among independent candidates. Second, each commits to that single result unconditionally, returning whatever the code produced, with no mechanism to recognise when no candidate is defensible and decline. The ability to abstain with a diagnosis, rather than emit a confident wrong answer, is a capability these single-trajectory pipelines simply lack.

Inference-time selection over multiple candidates (generate several solutions, keep the best) is among the most reliable accuracy levers known for code and reasoning, yet it  has received limited systematic study in agents answering closed-form questions over a dataframe. The reason is that the selection methods proven elsewhere lean on signals this setting does not provide: a single analytical question has no held-out validation metric and no test suite to score candidates against, and its result is a structured artefact (a named dataframe, sometimes with charts) rather than a scalar that answer-agreement voting could agree. Selection here must therefore be label-free, judge the executed artefacts directly, and, unlike majority voting, keep the option to abstain when no candidate is defensible. \Cref{sec:related} develops each of these requirements.

This paper presents \sysname{} (Generative, Ranked, Adaptive Code Ensemble), a production-oriented multi-agent pipeline that brings parallel best-of-$N$ selection to this label-free regime. \sysname{} dispatches $N$ diversified code-generation agents in parallel under a strict machine-checkable output contract, then selects among their executed results with a two-stage judge: a single-elimination pairwise tournament that pre-filters the pool, then a domain-specialised guardrail that either approves the best candidate or rejects the whole pool. On rejection it triggers an adaptive retry that re-diversifies the pool, routes a targeted critique to each agent, and forces every retry to add a new code invariant rather than a cosmetic edit. Our work is an integrated instantiation of these ingredients for the label-free dataframe-analysis regime: candidate selection is performed at the executed-artefact level and can abstain, rather than relying on scalar answer agreement or validation metrics. We make four contributions:

\begin{enumerate}[leftmargin=*]
  \item \textbf{Parallel best-of-$N$ code generation for label-free dataframe analysis} (\Cref{sec:codegen}): $N$ concurrent agents under a strict output contract, diversified by persona hints, per-retry persona rotation, and a two-regime temperature schedule, with a neutral anchor agent that always reflects the model's default behaviour. Parallel sampling itself is not our novelty. Our contribution is to adapt it to label-free dataframe analysis by enforcing an output contract and diversifying rollouts through personas, rotation, and temperature scheduling.
  \item \textbf{An integrated artefact-level selector with abstention}
  (Sections \ref{sec:pairwise} and \ref{sec:guardrail}): we compose an untrained pairwise tournament with a domain-specialised post-execution guardrail that judges the executed dataframe artefacts and their generating code, either selecting the strongest candidate or rejecting the pool. The contribution is not the pairwise-judge primitive itself, which we adopt from prior work showing that LLMs compare two candidates more reliably than they score or rank many at once~\cite{zheng2023judge,liu2025pairjudge,qin2024prp}, but its integration with dataframe artefact contracts, pool-level rejection, abstention, and targeted retry feedback in a label-free setting.
  \item \textbf{Adaptive retry with invariant-forcing feedback} (\Cref{sec:guardrail}): rejection re-diversifies the pool and routes per-agent critiques, with every code retry required to add a structural assertion rather than a surface fix.
  \item \textbf{A controlled $3{\times}3$ study with paired-bootstrap significance testing and equivalence analysis} (\Cref{sec:experiments,sec:results}): a paired-bootstrap analysis under Benjamini--Hochberg control that sweeps the number of parallel agents against the retry budget (jointly, where prior best-of-$N$ data and code work scales one axis and fixes the other), isolating a substitutability effect between the two levers and a named cost--accuracy operating point for deployment.
\end{enumerate}

% =====================================================================
\section{Background and Related Work}
\label{sec:related}

\paragraph{LLM-based data-analysis agents.}
A growing body of work casts tabular analysis as code generation over a
dataframe: the agent writes Python, executes it, and returns the result.
These code-writing agents~\cite{zhang2023datacopilot,hong2024datainterpreter,qiao2023taskweaver,sun2024lambda,you2025datawise}
differ in how they plan but build on standard agent primitives and
orchestration frameworks such as ReAct, Reflexion, and AutoGen~\cite{yao2023react,shinn2023reflexion,wu2023autogen},
and all share two limitations: each explores a single solution trajectory per question---iterating
or debugging within it, but never generating independent candidates to choose
among---and commits to its result unconditionally; the two assumptions \sysname{}
drops. A separate line of data-science automation does select among
candidates: SPIO, AIDE, and SELA~\cite{seo2025spio,jiang2025aide,chi2024sela}
run explicit generate--evaluate--select search, and DS-Agent~\cite{guo2024dsagent}
revises via case-based performance feedback. Either way they rank or refine
multi-step ML pipelines against a held-out validation metric---a
near-ground-truth signal that the single closed-form questions we target do not
provide, so our selection must instead be label-free.

\paragraph{Best-of-$N$ selection.}
Inference-time selection over multiple candidates is among the most reliable accuracy levers for LLMs: Self-Consistency~\cite{wang2022selfconsistency} majority-votes over sampled chains of thought, inference-scaling reports near-log-linear gains across hundreds of samples on code and math~\cite{brown2024monkeys,snell2024scaling}, and \emph{More Agents}~\cite{li2024moreagents} shows accuracy rising with sampled-agent count under voting---which our $N$-sweep revisits for dataframe code. Each of these moves a single lever, though: the parallel axis is scaled with no sequential-retry counterpart and no operating-point rule (\emph{More Agents} leaves cost optimisation to future work), while sequential self-refinement fixes the retry budget by hand~\cite{madaan2023selfrefine}. Only Snell et al.~\cite{snell2024scaling} sweep both jointly, but on MATH (a competition-mathematics reasoning benchmark) and with a difficulty-dependent, non-substitutable conclusion; our significance-controlled sweep of agents against retries has no counterpart for label-free dataframe code.

The selection step itself comes in several forms. Some train a verifier to score candidates~\cite{cobbe2021gsm8k,lightman2023verify,ni2023lever}; others need no training, either clustering programs by their execution output (CodeT~\cite{chen2022codet}, AlphaCode~\cite{li2022alphacode}) or prompting an LLM to judge the candidates directly (Universal Self-Consistency~\cite{chen2023usc}, Mixture-of-Agents~\cite{wang2024moa}), with the diversity of the candidate pool itself a further lever (SimpleStrat~\cite{wong2024simplestrat}). Text-to-SQL is where generate-and-select is already standard---CHASE-SQL~\cite{pourreza2024chasesql} trains a pairwise selector, PairJudge RM~\cite{liu2025pairjudge} runs a pairwise knockout, and MAC-SQL~\cite{wang2023macsql} coordinates several agents---yet even there the only \emph{label-free} selector is answer-agreement voting~\cite{wang2022selfconsistency}. None of these mechanisms has reached closed-form dataframe agents, where the two assumptions of \Cref{sec:introduction} persist: those agents~\cite{hong2024datainterpreter,sun2024lambda,you2025datawise} stay single-rollout, and voting cannot abstain. \sysname{} closes this gap by composing these ingredients into a label-free selector that judges executed artefacts.

\paragraph{LLM-as-judge, verification, and guardrails.}
LLM-as-judge approaches directly assign scores to candidate outputs~\cite{zheng2023judge,liu2023geval,kim2023prometheus}. A consistent finding is that LLMs are more reliable when comparing two candidates than when evaluating a single one in isolation~\cite{zheng2023judge} or ranking multiple candidates simultaneously~\cite{qin2024prp}. A related sequential paradigm is iterative self-refinement~\cite{madaan2023selfrefine,shinn2023reflexion,gou2023critic}. In parallel, guardrail frameworks such as NeMo~\cite{rebedea2023nemo} and Constitutional AI~\cite{bai2022constitutional} primarily focus on safety constraints rather than candidate selection.

\sysname{} departs from prior work on three fronts: its guardrail judges post-execution artefacts (the resulting named dataframe) rather than textual outputs, extending the execution signal of LEVER~\cite{ni2023lever} to a best-of-$N$ setting instead of a re-ranking pipeline; it combines the pairwise tournament with a guardrail that validates at most two finalists and can reject them outright, rather than relying on a single selection mechanism; and on rejection it regenerates the entire pool rather than refining a single output. Finally, unlike CHASE-SQL’s trained binary selector~\cite{pourreza2024chasesql}, our pairwise judge is prompt-based and needs no training, and its critiques for losing candidates double as retry feedback (\Cref{sec:guardrail}).

% =====================================================================
\section{The \sysname{} Pipeline}
\label{sec:method}

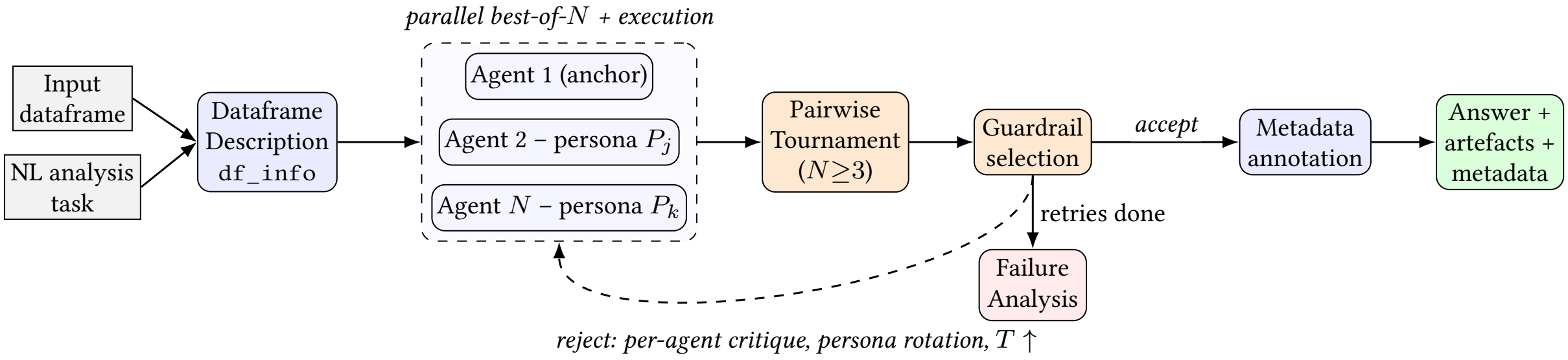
\begin{figure}[!ht]
\centering
\resizebox{\textwidth}{!}{%
\begin{tikzpicture}[
  font=\scriptsize,
  node distance=5mm and 7mm,
  io/.style={draw, align=center, minimum height=7mm, inner sep=2pt, fill=black!5},
  proc/.style={draw, rounded corners, align=center, minimum height=7mm, inner sep=2.5pt, fill=blue!8},
  agent/.style={draw, rounded corners, align=center, minimum width=20mm, minimum height=5mm, inner sep=1.5pt, fill=blue!4},
  sel/.style={draw, rounded corners, align=center, minimum height=7mm, inner sep=2.5pt, fill=orange!18},
  outbox/.style={draw, rounded corners, align=center, minimum height=7mm, inner sep=2.5pt, fill=green!14},
  arr/.style={-{Latex[length=1.8mm]}, semithick},
]
% inputs
\node[io] (df0) {Input\\dataframe};
\node[io, below=2.5mm of df0] (task) {NL analysis\\task};
% df description
\node[proc, right=of df0, yshift=-4.75mm] (desc) {Dataframe\\Description\\\texttt{df\_info}};
% agent pool -- the middle agent (a2) is aligned with desc, so the pool centre
% and the whole desc -> pool -> tournament -> guardrail -> output spine share
% one horizontal axis and the connecting arrows stay straight.
\node[agent, right=11mm of desc] (a2) {Agent 2 -- persona $P_j$};
\node[agent, above=2mm of a2] (a1) {Agent 1 (anchor)};
\node[agent, below=2mm of a2] (a3) {Agent $N$ -- persona $P_k$};
\begin{scope}[on background layer]
  \node[draw, dashed, rounded corners, fit=(a1)(a2)(a3), inner sep=3pt, fill=blue!2, label={[font=\scriptsize\itshape]above:parallel best-of-$N$ + execution}] (pool) {};
\end{scope}
% selection
\node[sel, right=of pool] (tour) {Pairwise\\Tournament\\($N{\geq}3$)};
\node[sel, right=of tour] (guard) {Guardrail\\selection};
% outputs
\node[proc, right=16mm of guard] (meta) {Metadata\\annotation};
\node[outbox, right=of meta] (final) {Answer +\\artefacts +\\metadata};
\node[proc, below=8mm of guard, fill=red!8] (fail) {Failure\\Analysis};
% arrows
\draw[arr] (df0.east) -- (desc.west);
\draw[arr] (task.east) -- (desc.west);
\draw[arr] (desc.east) -- (pool.west);
\draw[arr] (pool.east) -- (tour.west);
\draw[arr] (tour.east) -- (guard.west);
\draw[arr] (guard.east) -- (meta.west) node[pos=0.5, above, font=\scriptsize\itshape, fill=white, inner sep=1pt] {accept};
\draw[arr] (meta.east) -- (final.west);
\draw[arr] (guard.south) -- (fail.north) node[pos=0.5, right=1pt, font=\scriptsize, fill=white, inner sep=1pt] {retries done};
% retry feedback loop
\draw[arr, dashed] (guard.south) .. controls +(0,-12mm) and +(0,-12mm) .. (pool.south)
  node[pos=0.5, below=10pt, align=center, font=\scriptsize\itshape, fill=white, inner sep=1.5pt] {reject: per-agent critique, persona rotation, $T\uparrow$};
\end{tikzpicture}%
}
\caption{Detailed architecture of the \sysname{} pipeline. The controller first builds the shared dataframe description and dispatches $N$ parallel code-generation agents. Executed candidates that satisfy the output contract enter the selection stage: for $N{\geq}3$, a pairwise tournament reduces the pool to two finalists; otherwise, the tournament is bypassed. The guardrail then accepts one candidate or rejects the pool. Rejection triggers agent-specific feedback, persona rotation, and a new generation attempt until the retry budget $r$ is exhausted. Accepted artefacts receive metadata, whereas terminal failures produce a diagnostic report.}
\label{fig:pipeline}
\end{figure}

\sysname{} maps an input dataframe and a natural-language task to a single named result dataframe, optionally accompanied by chart artefacts when the task requires them, with lightweight user-facing metadata, with no human in the loop. The architecture is organised into three stages, detailed in \Cref{fig:pipeline}. First, a dataframe-description agent constructs the shared schema used by $N$ parallel code-generation agents. Second, the executed candidates pass through a selection stage composed of an optional pairwise tournament and a post-execution guardrail. Third, an accepted result is annotated with metadata, whereas a rejected pool triggers an adaptive retry or, after budget exhaustion, a terminal failure analysis. The central controller coordinates six agent roles while keeping candidate generation, selection, and retry feedback logically separated.

Throughout the paper, $N$ denotes the number of code-generation agents executed in parallel within each attempt, while $r$ denotes the maximum number of pool-level retries allowed after guardrail rejection. Thus, $r{=}0$ permits only the initial attempt, whereas $r{=}4$ permits the initial attempt followed by at most four retries.

\subsection{Dataframe Description}
\label{sec:dfdesc}

A dedicated agent previews the first twenty rows and emits a per-column semantic description (meaning, domain, role) merged with data types and sample values into a compact schema, \texttt{df\_info}. Computed once and reused by every downstream prompt, it is the sole authoritative description of the dataframe, keeping prompts compact and disambiguating similarly-named columns.

\subsection{Parallel Code Generation and Execution}
\label{sec:codegen}

A configurable number $N$ of code agents run concurrently, each interpreting the task independently, generating a self-contained Python script, and executing it in a sandbox over a fixed library set (NumPy, pandas, scikit-learn, PyTorch, statsmodels, Plotly). A per-agent timeout terminates stragglers without affecting the others, so the round is bounded by the slowest surviving agent. On an execution error (uncaught exception, missing result variable, or contract violation) an agent may retry internally to fix its own code before its slot is marked failed, repairing low-cost mistakes locally. Because this repair runs before selection, every candidate reaching the tournament or guardrail has already executed and satisfied the contract: the judges compare only runnable results, never broken code, and an agent that still fails or times out contributes no candidate rather than a faulty one.

A post-execution validator enforces a three-clause contract: (i)~a single-entry dictionary whose value is a pandas \texttt{DataFrame} (scalars, multi-table results, lists, and empty frames forbidden); (ii)~optional chart artefacts, when needed to answer the task, stored outside the result dataframe and linked back to it through lightweight provenance metadata; (iii)~a user-meaningful dataframe name (the dictionary key) for the metadata stage. Structural failures (scalar leakage, empty frames, dangling charts) thus become trivially detectable and, on our benchmarks, fall to near-zero, leaving semantic error as the dominant residual failure mode (\Cref{sec:results}).

\subsection{Diversification: Personas, Rotation, and Temperature}
\label{sec:diversification}

Agents are perturbed along three axes to reduce the risk that the pool collapses onto one interpretation. The motivating premise is a general one: across the generation-diversity literature, injecting variation into a generator---through sampling temperature, prompt-level stratification, or agent composition---helps it escape a single dominant mode and cover more of the solution space than repeated near-greedy decoding~\cite{wong2024simplestrat,sheng2025csc,wang2024moa}. We take this principle as motivated design and instantiate it as three concrete levers; we do not re-validate the principle here, nor do we claim each lever is independently supported, and ablations isolating their individual contributions are left to future work. \emph{(i)~Personas.} Each agent beyond the first carries one of four fixed preambles that bias its decisional stance without prescribing implementation (full text in \Cref{app:prompts}, item~A). Agent~1 carries no persona and serves as a neutral anchor, identical to the $N{=}1$ pipeline, so the model's most-likely answer is always in the pool. \emph{(ii)~Rotation.} At each retry the persona offset shifts by one, so an agent whose persona led it off the right track receives a different bias next time and the guardrail's per-agent critique (\Cref{sec:guardrail}) does not fight the same biased prior every iteration. \emph{(iii)~Temperature.} A single value, uniform across agents within an attempt, follows a two-regime schedule, where t denotes the attempt index, with $t=1$ indicating the initial attempt:
\begin{equation}
  T(t) = \begin{cases} 0.65 & t = 1 \\ 1.0 & t \geq 2 \end{cases}
  \label{eq:temp}
\end{equation}
The first regime favours clean, coherent code; the second escapes interpretation modes that per-agent advice cannot dislodge. Diversity is delegated to personas and rotation rather than temperature, since near-greedy temperatures starve selection of diversity while accuracy plateaus across a moderate-to-high band, and prompt-level stratification yields more useful diversity than temperature alone~\cite{wong2024simplestrat,sheng2025csc}.

\subsection{Pairwise Tournament}
\label{sec:pairwise}

For $N{\ge}3$ valid results a single-elimination bracket runs before the guardrail; for $N{<}3$ it is bypassed, recovering the no-tournament behaviour exactly. We adopt rather than propose it: that LLMs compare two candidates more reliably head-to-head than they rank many at once is an established finding we take from the literature and apply, not one we test~\cite{zheng2023judge,liu2025pairjudge,qin2024prp}. Each match compares two candidates (code plus a preview of the executed result dataframe) against the task and must enumerate concrete defects in both before declaring a winner~\cite{zheng2023judge}; the JSON verdict carries a \texttt{winner}, a \texttt{reason}, and a one-to-two-sentence \texttt{loser\_critique} stored per agent for routing on rejection. The bracket halts at two survivors, not one ($N{-}2$ judge calls), so both finalists keep an advice channel into the guardrail.

\subsection{Guardrail Validation and Adaptive Retry}
\label{sec:guardrail}

Because the tournament always precedes it (\Cref{sec:pairwise}), the guardrail receives at most two candidates---the two finalists when $N{\ge}2$, or the single candidate when $N{=}1$---each as its Python code paired with a preview of the executed result dataframe (so it judges the program and its artefact jointly, not the artefact alone), and emits a structured JSON verdict in one of two shapes.

\paragraph{Case A: at least one candidate is correct.}
It selects the candidate with the fewest and least severe defects and emits a user-facing narrative in business language, stripped of implementation detail and of any reference to the selection process, so the artefact is directly consumable by a front-end.

\paragraph{Case B: no candidate is correct.}
It emits a global root-cause Advice and a per-agent critique list (\texttt{Per\_agent\_advice}), one targeted entry per finalist it evaluated. This triggers a pool-level retry, up to $r$ attempts, that updates the pool at once: temperature steps (\Cref{eq:temp}), personas rotate (\Cref{sec:diversification}), global advice enters every prompt, and each per-agent critique reaches only its agent. The feedback channel routes by tournament outcome: the (up to two) finalists receive their \texttt{Per\_agent\_advice}, while each candidate eliminated earlier in the bracket (only when $N{\ge}3$) receives the \texttt{loser\_critique} from the match that knocked it out (global $\to$ all, per-agent $\to$ finalists, loser-critique $\to$ eliminated), so no agent retries on stale feedback. Every code retry must moreover add a new explicit invariant, assertion, or state variable addressing the root cause: patching only the symptom is forbidden, pushing retries toward structural rather than cosmetic fixes, on the hypothesis that the residual failure after one attempt is rarely fixable by a surface change.

\subsection{Failure Analysis and Metadata}
\label{sec:metadata}

Two terminal roles close the pipeline. If the retry budget is exhausted with no approval, a failure-analysis agent returns a diagnostic narrative (what each agent attempted, the root cause, the evolution across retries) in place of a result, so an unrecoverable case degrades into actionable feedback rather than a silent error. On success, a metadata agent adds concise titles and descriptions to the accepted dataframe and to any optional chart artefacts. These annotations are retained for front-end consumption, but they are not part of the experimental metric.

% =====================================================================
\section{Experimental Setup}
\label{sec:experiments}

We evaluate on DAEval (InfiAgent-DABench~\cite{hu2024infiagent}), closed-form data-analysis questions over CSV files with structured expected answers. The metric is \emph{answer match}: correct if all expected values match, under a $1\%$ relative / $0.01$ absolute numeric tolerance and case-insensitive string comparison. Each prediction is a match or a wrong response (structurally valid but not matching gold) with structural failures to near-zero due to the strict output contract. The evaluation therefore targets the analytical answer, not presentation quality.

The pipeline uses the Google Agent Development Kit with LiteLLM for model-agnostic access and Langfuse for observability; code runs in a sandbox over the library set of \Cref{sec:codegen}, with artefacts on S3-compatible storage. Our primary backbone, shared by all agents and the judge/guardrail, is \mbox{Qwen3.6-35B-A3B}~\cite{qwen2026}, a general-purpose mixture-of-experts model with strong coding ability ($35$B total / $3$B active per token, $8$ routed plus $1$ shared of $256$ experts, $262{,}144$-token native context), served at full precision (BF16); unless stated otherwise, all reported results use this backbone. To probe how the pipeline degrades under a smaller, dense model, we additionally run the DAEval $3{\times}3$ sweep with \mbox{Qwen3.5-9B}~\cite{qwen3.5-9b} ($9$B dense parameters) under an otherwise identical configuration. The purpose of the 9B experiment is therefore not to provide a complete cross-benchmark model comparison, but to test whether the interaction between agents and retries persists under a smaller backbone. DAEval was selected for this probe because it is the primary and more analytically demanding benchmark in our evaluation. Extending the full $3{\times}3$ grid to DataBench would require a further $3{,}600$ question-level executions, in addition to retries and judging calls, and was outside the computational scope of this study. Token usage and wall-clock time are measured per question via Langfuse; the full agent and persona prompts are reproduced in Appendix \ref{app:prompts}.

The experiments were orchestrated from a workstation with 13\,GiB of memory and Python 3.12.3. Both Qwen models were hosted internally by Engineering Ingegneria Informatica S.p.A., served on a single NVIDIA H100 GPU, and accessed through LiteLLM. All sweep cells used the same remote infrastructure and model settings. Agents were submitted concurrently, and token usage and end-to-end latency were recorded through Langfuse.

We sweep a $3{\times}3$ grid over two levers: the number of parallel agents (the pool size $N \in \{1,2,4\}$) and the retry budget $r \in \{0,4,10\}$ (the pool-level guardrail-rejection rounds of \Cref{sec:guardrail}). The range is deliberately small: each \sysname{} agent is a full code-generation, execution, and judging rollout, not a token-cheap sample, so the question is how little of each lever serves. We use the $M{=}246$ DAEval questions with a well-formed, machine-checkable gold answer (the rest of the 257-question set have malformed or missing values, dropped uniformly across cells), so every configuration sees the identical question set in identical order, paired at the question level. The cheapest cell $(N{=}1, r{=}0)$ is the minimal single-rollout baseline. For benchmark context, InfiAgent-DABench reports a GPT-4 baseline of $78.99\%$ on DAEval~\cite{hu2024infiagent}. This value is provided only as an external reference because the backbone and evaluation harness differ from ours, preventing a controlled system-level comparison.

The comparison between this minimal baseline and the remaining sweep cells is intentionally an inference-scaling comparison rather than a compute-matched architectural ablation: increasing $N$ or $r$ necessarily increases the opportunity to generate a correct candidate. Consequently, the baseline-to-best gain measures the practical end-to-end benefit of allocating additional inference compute to the complete pipeline, but does not by itself establish that the same budget would be used less effectively by repeated independent sampling or generic sequential self-refinement. To partially separate candidate generation from selection without additional model runs, \Cref{sec:guardrail-quality} compares the tournament+guardrail against majority answer-agreement voting and random selection on exactly the same candidate pools. These selectors therefore receive identical generated candidates and share the same candidate-generation budget, although the tournament+guardrail incurs additional judging calls and is not compute-matched in total inference cost.

We repeat the identical $3{\times}3$ sweep, backbone, and metric on a second, independent benchmark, DataBench~\cite{oses2024databench}, a large-scale QA corpus over real-world tables spanning five domains, with typed closed-form answers (boolean, category, number, lists). The two probe different question styles: DAEval targets data-analysis concepts from expert interviews---summary statistics, distribution and correlation analysis, outlier detection, preprocessing, feature engineering, machine learning~\cite{hu2024infiagent}, often requiring a multi-step statistical or modelling computation---whereas DataBench poses fact-oriented questions answerable by simpler column operations (counting, filtering, aggregation, top-$k$ ranking, threshold comparisons), with correlation and machine learning largely absent. This difficulty gap partly explains DataBench's far higher single-rollout baseline (\Cref{sec:results}). Representative questions from both benchmarks are reported in \Cref{tab:dataset-examples}.
\begin{table}[!ht] \centering \caption{Representative questions from the two evaluation benchmarks. DAEval includes multi-step statistical and modelling tasks, whereas DataBench mainly targets direct retrieval, aggregation, and ranking over tabular data.} \label{tab:dataset-examples} \small \begin{tabularx}{\textwidth}{@{}lX@{}} \toprule \textbf{Benchmark} & \textbf{Representative questions} \\ \midrule DAEval & \begin{minipage}[t]{\linewidth} \begin{itemize}[leftmargin=*, nosep] \item Calculate the mean, median, and population standard deviation of passenger fares separately for each passenger class, and interpret the results. \item Encode categorical variables and train and evaluate a regression model to predict passenger survival. \end{itemize} \end{minipage} \\ \addlinespace DataBench & \begin{minipage}[t]{\linewidth} \begin{itemize}[leftmargin=*, nosep] \item How many billionaires are there in the Technology category? \item List the three countries with the most billionaires. \end{itemize} \end{minipage} \\ \bottomrule \end{tabularx} \end{table}
To bound cost we draw a fixed subset of $M'{=}400$ questions ($80$ tables $\times$ $5$, seed $12345$) from the full $1{,}829$, stratified by file; the answer types come out near-balanced (boolean $79$, category $75$, number $92$, list[number] $81$, list[category] $73$) and the subset is held constant and ordered across all nine cells, as for DAEval. For external context, the SemEval-2025 Task~8 leaderboard reports a best score of $95.0\%$, a best proprietary-system score of $89.9\%$, and several competitive systems in the mid-to-high $80\%$ range~\cite{oses2025semeval}. These scores were obtained on the full test set with different models and harnesses. They are therefore reported as benchmark references rather than as directly comparable baselines.

We estimate match-rate differences with a paired bootstrap over the common question set, using $10{,}000$ resamples and percentile confidence intervals. Multiple comparisons are controlled with the Benjamini--Hochberg procedure at $q{=}0.1$. Significance symbols report adjusted values (* $p_\text{adj}{<}0.1$, ** ${<}0.05$, *** ${<}0.01$). Token usage and wall-clock time are normalised to the median of the single-rollout baseline.

For the four plateau comparisons underlying the substitutability claim, we additionally test statistical equivalence using two one-sided tests. Equivalence is established when the $90\%$ paired interval lies within a benchmark-specific margin equal to the median measurement resolution: $\pm5.7$\,pp for DAEval and $\pm3.75$\,pp for DataBench. This prevents a non-significant difference from being interpreted automatically as evidence of equal performance.

% =====================================================================
\section{Results}
\label{sec:results}

\begin{figure}[!ht]
  \centering
  \begin{subfigure}[b]{0.85\textwidth}
    \centering
    \includegraphics[width=\linewidth]{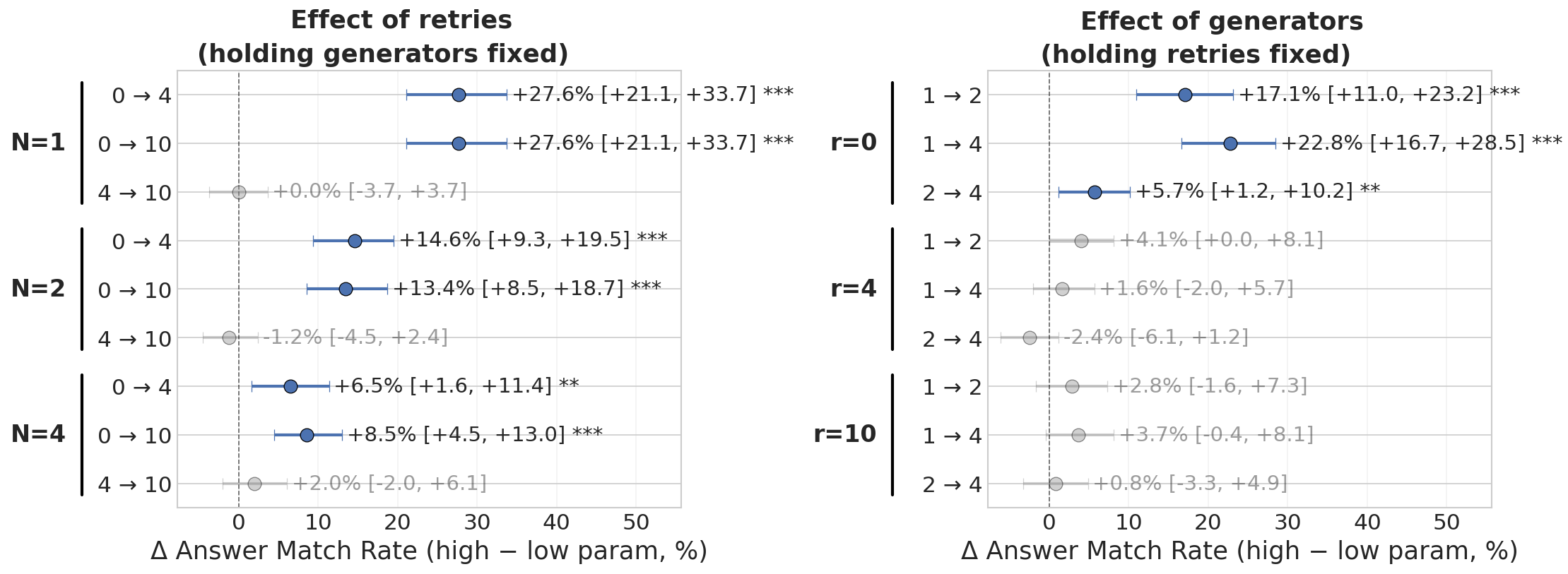}
    \caption{Qwen3.6-35B-A3B (MoE) backbone.}
    \label{fig:pairwise}
  \end{subfigure}\\[0.2em]
  \begin{subfigure}[b]{0.85\textwidth}
    \centering
    \includegraphics[width=\linewidth]{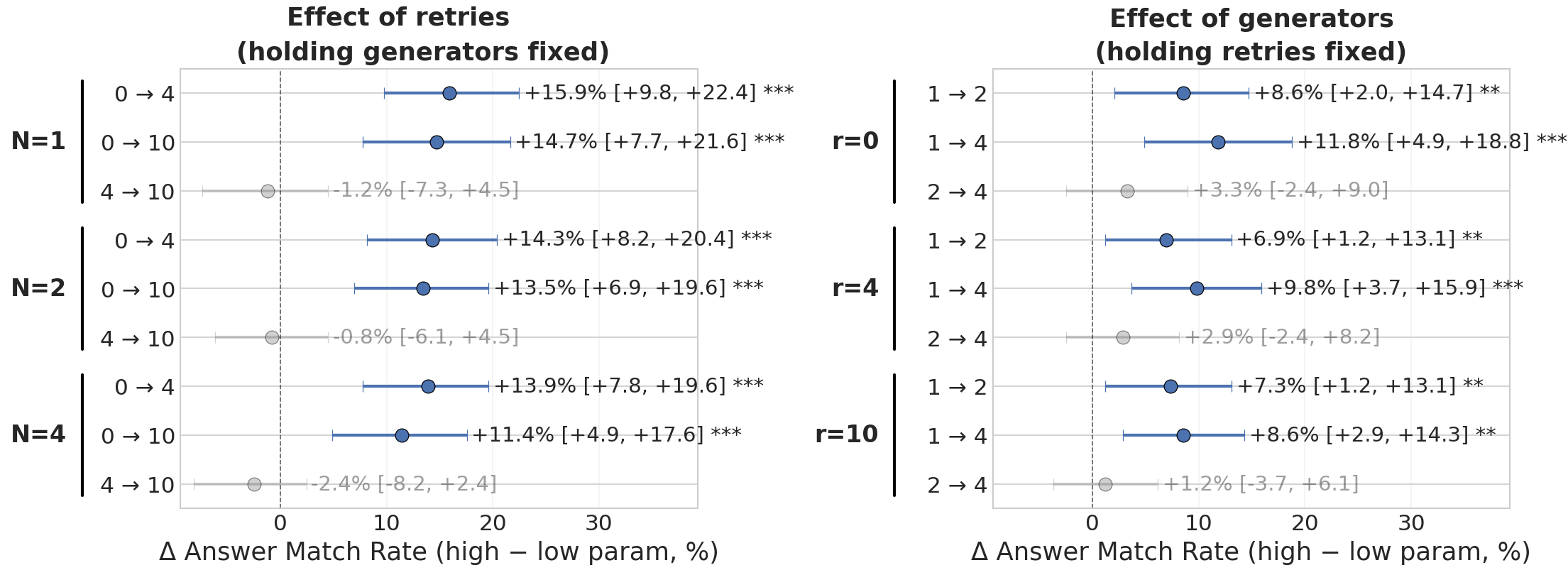}
    \caption{Qwen3.5-9B (dense) backbone.}
    \label{fig:pairwise-9b}
  \end{subfigure}
  \caption{Single-factor simple effects on DAEval ($M{=}246$) under the paired-bootstrap + BH protocol of \Cref{sec:experiments}: retries at fixed $N$ (left) and agents at fixed $r$ (right). (a)~With the primary 35B-A3B MoE backbone, retries are decisive only at $N{=}1$ and agents only at $r{=}0$, with nothing significant beyond one step of either lever. (b)~The same view for the smaller Qwen3.5-9B backbone (downscaling probe). Significance stars denote the Benjamini--Hochberg-adjusted $p$-value (* $p_\text{adj}{<}0.1$, ** ${<}0.05$, *** ${<}0.01$; n.s.\ otherwise).}
  \label{fig:sweep}
\end{figure}

\paragraph{DAEval with the 35B backbone.} The complete pipeline improves answer match from \keyresult{$39\%$} for the single-rollout baseline to a maximum of \keyresult{$71\%$} (\Cref{tab:sweep}). With no retries, increasing the number of agents raises accuracy substantially, reaching approximately $62\%$ with four agents. A retry budget produces a larger improvement for the single-agent configuration and brings all configurations into a narrow high-accuracy region. Every non-baseline configuration significantly outperforms the baseline after correction for multiple comparisons. At the question level, the best configuration recovers more than half of the answers missed by the baseline while preserving almost all previously correct answers. The simple-effect analysis in \Cref{fig:pairwise} shows that agents and retries are largely substitutable for this backbone: agents are most useful when retries are unavailable, whereas retries are most useful with a single agent. Once either resource is sufficiently available, increasing the other produces only minor differences. The equivalence analysis supports this interpretation for all four plateau comparisons.

\paragraph{DAEval with the 9B backbone.} The smaller dense backbone exhibits the same overall benefit at a lower final accuracy. The single-rollout baseline is comparable to that of the 35B model, but the gap between the backbones emerges when additional agents and retries are introduced. The dense model reaches a maximum answer match of \keyresult{$63\%$}, compared with \keyresult{$71\%$} for the MoE backbone. Retries again provide most of their benefit by the first tested retry level. Unlike the 35B model, however, the 9B model continues to benefit from additional parallel agents even when retries are available. This suggests that the smaller model gains more from independent interpretations of the question, whereas the stronger backbone can more often correct an initial interpretation through retry feedback. Detailed cell-level results are reported in \Cref{tab:sweep} and \Cref{fig:pairwise-9b}.

\paragraph{DataBench with the 35B backbone.} DataBench independently reproduces the main pattern despite its substantially higher single-rollout baseline. Answer match rises from approximately \keyresult{$70\%$} to \keyresult{$85\%$}, and all non-baseline configurations significantly outperform the baseline. As on DAEval, retries are most useful with a single agent, while additional agents contribute mainly when retries are unavailable. Performance above the baseline is concentrated within a narrow range, making the inexpensive single-agent configuration with a moderate retry budget nearly as accurate as the most expensive configuration. Equivalence is established for two of the four plateau comparisons; the remaining differences are small but cannot be declared equivalent under the stricter DataBench margin. The complete results are reported in \Cref{tab:sweep}.

The high-accuracy plateau also defines a clear cost--accuracy frontier. On both benchmarks, a single agent with a moderate retry budget approaches the best accuracy at approximately $1.1{\times}$ the baseline token cost. On DAEval, two agents with the same retry budget attain the highest accuracy at approximately twice the baseline cost. Four-agent configurations consume substantially more tokens without a corresponding improvement, while latency remains close to the baseline because agents execute concurrently. Thus, the practical trade-off is driven primarily by token usage rather than wall-clock time.

\begin{table}[H]
\centering
\caption{Answer-match rate across the $3{\times}3$ sweep, where $N$ is the number of parallel code-generation agents per attempt and $r$ is the maximum number of pool-level retries after guardrail rejection. For each backbone and benchmark, the table also reports the paired transition from the single-rollout baseline to the best-performing cell; baseline cell $(N{=}1,r{=}0)$ and best cell in bold. Top: DAEval ($M{=}246$) under the primary 35B-A3B MoE backbone; middle: DAEval under the smaller dense Qwen3.5-9B backbone (downscaling probe, \Cref{sec:experiments}); bottom: DataBench ($M'{=}400$, 35B-A3B). Single-factor simple effects for DAEval are in \Cref{fig:sweep}.}
\label{tab:sweep}
\small

\textbf{DAEval ($M{=}246$) --- Qwen3.6-35B-A3B (MoE)}\\[3pt]
\begin{tabular}{@{}c@{\hspace{1.5em}}|@{\hspace{1.5em}}l@{}}
\begin{tabular}[t]{@{}lccc@{}}
\toprule
\textbf{match rate (\%)} & $r{=}0$ & $r{=}4$ & $r{=}10$ \\
\midrule
$N{=}1$ & \textbf{39.0} & 66.7 & 66.7 \\
$N{=}2$ & 56.1 & \textbf{70.7} & 69.5 \\
$N{=}4$ & 61.8 & 68.3 & 70.3 \\
\bottomrule
\end{tabular}
&
\begin{tabular}[t]{@{}ll@{}}
\toprule
\multicolumn{2}{@{}l}{\textbf{baseline$\,\to\,$best} $(N{=}2,r{=}4)$, $246$ paired} \\
\midrule
\quad from correct ($96$) & $95$ kept, $1$ regress \\
\quad from wrong ($150$)  & $79$ recovered \ (net $+78$) \\
\bottomrule
\end{tabular}
\end{tabular}

\vspace{0.2em}
\textbf{DAEval ($M{=}246$) --- Qwen3.5-9B (dense)}\\[3pt]
\begin{tabular}{@{}c@{\hspace{1.5em}}|@{\hspace{1.5em}}l@{}}
\begin{tabular}[t]{@{}lccc@{}}
\toprule
\textbf{match rate (\%)} & $r{=}0$ & $r{=}4$ & $r{=}10$ \\
\midrule
$N{=}1$ & \textbf{37.8} & 53.5 & 52.0 \\
$N{=}2$ & 45.9 & 60.2 & 59.4 \\
$N{=}4$ & 49.2 & \textbf{63.4} & 60.6 \\
\bottomrule
\end{tabular}
&
\begin{tabular}[t]{@{}ll@{}}
\toprule
\multicolumn{2}{@{}l}{\textbf{baseline$\,\to\,$best} $(N{=}4,r{=}4)$, $246$ paired} \\
\midrule
\quad from correct ($93$) & $82$ kept, $11$ regress \\
\quad from wrong ($153$)  & $74$ recovered \ (net $+63$) \\
\bottomrule
\end{tabular}
\end{tabular}

\vspace{0.2em}
\textbf{DataBench ($M'{=}400$) --- Qwen3.6-35B-A3B (MoE)}\\[3pt]
\begin{tabular}{@{}c@{\hspace{1.5em}}|@{\hspace{1.5em}}l@{}}
\begin{tabular}[t]{@{}lccc@{}}
\toprule
\textbf{match rate (\%)} & $r{=}0$ & $r{=}4$ & $r{=}10$ \\
\midrule
$N{=}1$ & \textbf{69.8} & 84.5 & 83.5 \\
$N{=}2$ & 79.0 & 82.5 & 83.0 \\
$N{=}4$ & 78.5 & 82.2 & \textbf{85.0} \\
\bottomrule
\end{tabular}
&
\begin{tabular}[t]{@{}ll@{}}
\toprule
\multicolumn{2}{@{}l}{\textbf{baseline$\,\to\,$best} $(N{=}4,r{=}10)$, $400$ paired} \\
\midrule
\quad from correct ($279$) & $270$ kept, $9$ regress \\
\quad from wrong ($121$)   & $70$ recovered \ (net $+61$) \\
\bottomrule
\end{tabular}
\end{tabular}
\end{table}

\subsection{Selection Quality}
\label{sec:guardrail-quality}

End-to-end accuracy cannot distinguish coverage failures, where no correct candidate is generated, from selection failures. We therefore instrument a separate DAEval run at $(N{=}4,r{=}4)$ with the 35B backbone and grade every candidate against gold. Its $246$ questions produce $263$ selection decisions because retries re-enter selection. Each pool contains on average $3.66$ valid candidates, $2.32$ of which are correct, but candidates rarely produce equivalent artefacts.

\paragraph{Comparison with voting.}
To control for the main source of inference scaling, we compare three selectors on exactly the same $263$ candidate pools (\Cref{tab:guardrail}). All methods therefore receive identical generated candidates and have the same candidate-generation budget; only the selection mechanism changes. Candidates are clustered by output equivalence (matching type, shape, columns, and preview) so that answer-agreement voting has a well-defined input. The tournament+guardrail is correct on \keyresult{$66.9\%$} of decisions---\keyresult{$86\%$} of the way to the \keyresult{$77.6\%$} oracle ceiling. It significantly beats majority answer-agreement voting (\keyresult{$63.1\%$}; \keyresult{$+3.8$\,pp}, paired-bootstrap $95\%$\,CI $[+0.3,+7.3]$, $p{=}0.03$) and a random pick (\keyresult{$62.9\%$}; \keyresult{$+4.0$\,pp}, $p{=}0.02$). Thus, additional candidate generation alone does not explain the complete benefit: holding the generated pool fixed, the proposed selector extracts a modest but statistically significant improvement over voting. This is a controlled comparison of selection quality, not a claim of equal total inference cost, because the tournament and guardrail require additional judge calls.

Voting, by contrast, is statistically no better than random ($+0.2$\,pp, $p{=}0.66$): the candidates rarely coincide, so there is little agreement to count---in \keyresult{$63.9\%$} of decisions no two agree at all, and a majority cluster of size ${\geq}2$ exists in only $34.2\%$. The pools are already majority-correct, so a random pick alone secures $62.9\%$.

\paragraph{Selection and abstention.} 
Conditioning on whether the pool contains a correct candidate reveals two distinct behaviours (\Cref{tab:guardrail}). Selection is the stronger one: when a correct candidate is available, the selector chooses it in \keyresult{$86.3\%$} of cases. Most of the remaining oracle gap is caused either by choosing an incorrect candidate or by rejecting a pool that already contains a valid answer. Abstention is considerably weaker. When every candidate is incorrect, the guardrail rejects the pool in only \keyresult{$10.2\%$} of cases and usually accepts an incorrect answer. The system is therefore better at ranking available candidates than at recognising that none is defensible.

\begin{table}[H]
\centering
\caption{Selection quality on DAEval ($N{=}4$, $r{=}4$; $263$ decisions), each conditioned on whether the pool held a correct candidate (within-row \% in parentheses). Realized selection ($176/263{=}66.9\%$ correct) is bounded by the oracle ceiling ($204/263{=}77.6\%$, a correct candidate present) and beats the label-free baselines majority answer-agreement voting ($63.1\%$) and random pick ($62.9\%$), which are statistically indistinguishable.}
\label{tab:guardrail}
\small
\begin{tabular}{@{}lcccc@{}}
\toprule
\textbf{correct candidate in pool?} & \textbf{sel.\ correct} & \textbf{sel.\ wrong} & \textbf{rejected} & \textbf{total} \\
\midrule
yes & \textbf{176 (86.3\%)} & 17 (8.3\%) & 11 (5.4\%) & 204 \\
no  & --                   & 53 (89.8\%) & 6 (10.2\%) & 59 \\
\midrule
all & 176 (66.9\%) & 70 & 17 & 263 \\
\bottomrule
\end{tabular}
\end{table}

\paragraph{Retry after rejection.}
Rejected pools trigger a re-diversified retry. Almost half of the affected questions eventually reach a correct final answer, although recovery is more frequent when the rejected pool already contained a valid candidate. These values are based on a small number of rejected questions and should therefore be interpreted qualitatively. Retry also explains why per-question accuracy exceeds per-decision selection accuracy. A rejected decision counts as unsuccessful in the selection analysis but can still produce a correct answer in a later attempt.

% =====================================================================
\section{Discussion}
\label{sec:discussion}

The central finding, on the 35B backbone, is that the two ways of spending more compute---running more agents in parallel ($N$) or allowing more retries ($r$)---are largely interchangeable: once one of them is generous, adding the other barely helps (\Cref{fig:pairwise}; the flat $r{\geq}4$ plateau in \Cref{tab:sweep}). A plausible explanation is that the remaining errors are mostly semantic misreadings after structural failures have been filtered by the output contract. The output contract removes almost all structural failures, so what is left is the model misreading the question---and for that kind of error a failed attempt is still useful, because it shows what went wrong and lets the next try, whether a fresh parallel agent or a retry, start better informed. Parallel agents and retries are thus two ways of taking more shots at the same target, which is why having plenty of one makes the other redundant. This agrees with \emph{More Agents}~\cite{li2024moreagents} and differs from Snell et al.~\cite{snell2024scaling}, where the best split between parallel and sequential compute depends on problem difficulty. Consistent with this, the loop recovers \keyresult{$53\%$} of initially wrong answers (\Cref{sec:results}): high for a fully automatic system, plausibly because executing the generated Python yields concrete feedback and each retry is forced to add a new assertion rather than a cosmetic patch (\Cref{sec:guardrail}). The weaker 9B backbone sharpens the picture: there parallel agents keep helping even under a generous retry budget (\Cref{sec:results}), because a smaller model's single reading is less often the right one, so the extra interpretations that more agents supply carry a benefit retries cannot---whereas at 35B, whose first reading is usually right, that benefit has largely vanished and the two levers collapse into substitutes.

The pattern is not a single-benchmark artefact: it survives DataBench's far higher baseline and halved headroom (\Cref{sec:results}). Substitutability still holds on the plateau, even if added retries now buy a small but significant increment at $N{\ge}2$ rather than ceasing to help past $N{=}1$. One caveat tempers the cross-benchmark comparison: DataBench's high baseline is partly a floor effect of typed answers---driven by its categorical questions rather than its booleans, which are in fact among the hardest types here---so it overestimates the true difficulty gap.

For deployment, this substitutability together with the cost-efficiency frontier (\Cref{sec:results}) points to $(N{=}2, r{=}4)$ as the operating point for standard workloads and $(N{=}1, r{=}4)$ as a budget fallback; at these pool sizes ($N{<}3$) the tournament is bypassed (\Cref{sec:pairwise}), so deployment leans on the guardrail alone. This is an operating point, not a ceiling: our questions are closed-form over clean single CSVs, and we cannot exclude that more complex dataframes or harder, more open-ended questions---where a single reading is less likely to be right---would reward the broader interpretive diversity of $N{=}4$, exactly the regime in which the pairwise tournament also becomes active.

The selection analysis (\Cref{sec:guardrail-quality}) shows the two links are unequal. Selection is the strong one---a correct candidate, when present, is usually recovered despite the pool rarely agreeing on it, which is why the tournament+guardrail clears a voting baseline that the unscored artefacts leave no better than random. Abstention is the weak link: the guardrail tells good candidates from bad yet seldom recognises when all are wrong, so it mostly false-accepts all-wrong pools. This weak abstention directly limits accuracy: all-wrong pools are usually false-accepted, whereas correctly rejecting them would trigger a re-diversified retry and create another opportunity for recovery. Rejections of pools containing a correct candidate are often recovered on retry, while genuinely all-wrong pools rarely turn around. Abstention is thus the axis to sharpen next---a capability the single-rollout and voting baselines lack entirely---and, the residual failure being semantic, the next gains lie in the guardrail/retry layer rather than in adding generators.

Four limitations bound these results. First, DAEval supplies each question's source CSV as a fixed, authoritative input. The data itself can be messy---missing values, outliers, and preprocessing are part of the task---but the table is the correct one for the question, so these numbers isolate the analytical stage. In deployment the dataframe is instead a fallible upstream output (a wrong table, or one carrying extraction or ETL errors), which compounds an error rate this benchmark does not exercise. End-to-end evaluation with imperfect upstream data is future work.

Second, the baseline and multi-agent/retry configurations are not compute-matched. Their comparison measures the practical benefit of inference scaling through the complete pipeline, but does not establish superiority over repeated sampling or generic self-refinement under an equal total budget. The same-pool analysis partially controls this confound: given identical candidates, tournament+guardrail selection outperforms majority voting by $3.8$\,pp (\Cref{sec:guardrail-quality}), although it still incurs additional judge calls. A fully compute-matched comparison requires new executions and is left to future work.

Third, the sweep does not isolate the marginal impact of personas, rotation, temperature, invariant forcing, or the pairwise tournament, which we adopt from prior work showing that LLMs judge pairwise comparisons more reliably than absolute scores or full rankings~\cite{zheng2023judge,liu2025pairjudge,qin2024prp}, rather than advancing it as an independently measured contribution. A per-component ablation that isolates each lever is therefore also left to future work.

Finally, answer match treats all errors uniformly and cannot distinguish near-misses from semantically incorrect answers or complete failures. A future qualitative error analysis should separate failures in interpretation, computation, formatting, and abstention.

% =====================================================================
\section{Conclusion}
\label{sec:conclusion}

\sysname{} addresses a gap between single-rollout dataframe agents, validation-metric pipeline selectors, and scalar-voting tabular-QA ensembles by making best-of-$N$ selection label-free, artefact-level, and able to abstain: diversified parallel agents under a strict contract, a pairwise tournament composed with a post-execution guardrail, and an adaptive retry that forces structural rather than cosmetic fixes. On DAEval, with the 35B-A3B backbone, increasing inference compute through the complete pipeline lifts answer match from \keyresult{$39\%$} to \keyresult{$71\%$} and recovers \keyresult{$53\%$} of previously wrong answers. A controlled same-pool comparison further shows that, for identical generated candidates, tournament+guardrail selection improves over majority answer-agreement voting by \keyresult{$3.8$\,pp}. The sweep also exposes a substitutability between agents and retries that identifies an efficient operating point. DataBench reproduces the same directional plateau pattern (\keyresult{$+15$\,pp}, \keyresult{$70\!\to\!85\%$}), with equivalence partially established there (two of four plateau steps clear TOST). A smaller dense backbone, Qwen3.5-9B, serves as a downscaling probe on DAEval, where the same pattern holds at a lower level (\keyresult{$+25$\,pp}, \keyresult{$38\!\to\!63\%$}): retries again saturate at $r{=}4$, while---unlike at 35B---added parallel agents keep helping even under a generous retry budget.

\begin{acknowledgments}
  This work is framed within the Project ``AVANT'' -- Project no. IPCEI-CL\_0000005 -- Application protocol no. 108421 of 14/05/2024 -- CUP B89J24002920005 -- Grant decree no. 1322 of August 8, 2024 -- financed by the European Union -- NextGenerationEU (IPCEI Funding).
\end{acknowledgments}

%% Mandatory for CEUR-WS since Jan 2025 -- see https://ceur-ws.org/genai-tax.html
\section*{Declaration on Generative AI}
During the preparation of this work, the authors used Microsoft Copilot for grammar checking, spelling correction, and text rephrasing. The authors subsequently reviewed and edited all generated suggestions and take full responsibility for the content of the manuscript.

% =====================================================================

% =====================================================================

\appendix

\section{Agent Personas and Prompt Templates}
\label{app:prompts}
We reproduce the four personas and three prompt templates (code/ML agent, pairwise judge, post-execution guardrail). \texttt{\{braces\}} are runtime placeholders; \texttt{>=} replaces the original ``$\geq$'' glyph.

{\scriptsize
\begin{multicols}{2}
\noindent\textbf{A. Personas ($P_1$--$P_4$).} One preamble is prepended to every
non-anchor agent (Agent~1, the anchor, carries none); the offset rotates by one
each retry (\Cref{sec:diversification}).

\noindent\emph{$P_1$, schema-strict / minimal.} Treat the provided
\texttt{df\_info} as the only definitive source of truth about the dataframe.
Use only columns and dtypes that are explicitly listed; never infer the
existence of fields, categories, or relationships that are not stated. Prefer
the smallest, most direct code that answers the question without adding checks
beyond what is required.

\noindent\emph{$P_2$, defensive / robust.} Be explicit about edge cases:
handle potential NaN values, verify dtypes before operations that assume them,
guard against empty groups or single-class targets, and prefer explicit
conditions over implicit defaults. Add assertions or state variables that make
assumptions about the data visible in the code.

\noindent\emph{$P_3$, exploratory / chart-generous.} When the question
implies inspecting a group of similar variables, produce the full set of plots
(one per variable), not just an example. When in doubt about the right grain,
lean toward keeping the result at the individual-row level with derived columns
added, rather than aggregating early.

\noindent\emph{$P_4$, pragmatic / direct.} Go as directly as possible to what
the user question literally asks. Avoid pre-processing, feature engineering, or
aggregations that the question does not require. Pick the simplest aggregation
or operation that matches the wording of the request.

\medskip
\noindent\textbf{B. Code/ML agent.} Prepended for agents $2..N$ with a line
|PERSONA: <one of P1..P4>|.
\begingroup
\scriptsize

You are an execution-only ML agent, given a dataframe at
\texttt{\{df\_path\}} with schema \texttt{\{df\_info\}}. Load it with the
provided \texttt{load\_df} helper, never \texttt{pd.read\_csv}; use only schema
columns---referencing non-existing columns is a semantic error.

Generate self-contained Python and execute it through
\texttt{execute\_generated\_code}; do not print, output raw code, or add
explanations outside the executed analysis. Reason only over the current
request, code, error, and provided retry context.

Handle missing values explicitly before modelling, transforming, aggregating,
or asserting. Critical operations (loading, fitting, transforming, predicting)
must expose success or failure through explicit state, assertions, or
control-flow checks. Silent exception handling is forbidden: a caught exception
must record its cause, and downstream operations must verify the required state
exists.

After execution, follow one of two paths. If the output contains
\texttt{"COMPLETED": true}, return only a business-language summary of the
findings. Otherwise treat the run as failed: identify the earliest causal
failure and the violated implicit assumption, then revise the code to enforce
that assumption explicitly. Each retry must add at least one new invariant,
assertion, or state variable for the root cause; cosmetic or trivially
equivalent fixes are invalid.

The script must define exactly one result dictionary entry:
\begin{quote}
\scriptsize\ttfamily
result = \{ "\textless descriptive\_name\textgreater": \{
\hspace*{1.5em}"type": "dataframe", "value": \textless pandas.DataFrame\textgreater
\} \}
\end{quote}
Only \texttt{"type": "dataframe"} is allowed; scalars, lists, dicts, plots,
metrics, and empty frames are forbidden (even scalar answers must be wrapped in
a non-empty dataframe). Consolidate multiple tables into one, kept at the finest
grain the answer or any chart needs.

When charts are required, save them only inside
\texttt{\{agent\_output\_dir\_\textless i\textgreater\}} via
\texttt{os.path.join}, never inside \texttt{result}, each generated from the
dataframe in \texttt{result}; produce the full set when the question needs one
chart per variable. For each Plotly chart save the triplet
\texttt{.plotly.json}, \texttt{.png}, \texttt{.recipe.json}, whose recipe field
\texttt{data\_csv} must match the single key in \texttt{result}.

For ML operations, calling \texttt{predict}, \texttt{transform}, or accessing
learned attributes such as \texttt{coef\_} or \texttt{classes\_} requires
explicit proof that fitting completed. If fitting fails or is skipped,
downstream ML operations must not run.

Retry feedback, when present, is provided through:
\texttt{\{retry\_advice\}} for global guardrail advice,
\texttt{\{retry\_advice\_\textless i\textgreater\}} for this agent's critique,
\texttt{\{error\_\textless i\textgreater\}} for the previous error, and
\texttt{\{python\_code\_\textless i\textgreater\}} for the previous code.

User request: \texttt{"\{ML\_question\}"}.

\endgroup
\noindent\textbf{C. Pairwise judge.}
\begingroup
\scriptsize

The pairwise judge receives a natural-language question
\texttt{\{ML\_question\}} and two executed candidate analyses,
\texttt{\{pairwise\_A\_ML\}} and \texttt{\{pairwise\_B\_ML\}}, including both
their Python code and produced results. Its task is to select the candidate
that better answers the question.

Before deciding, the judge internally critiques both candidates. It checks
whether each candidate uses only existing dataframe columns, handles missing
values before modelling or aggregation, chooses the correct metric or
aggregation, returns a \texttt{result} dataframe at the required grain, produces
all required plots, and fully answers the question rather than only part of it.
Semantic defects such as wrong columns, wrong aggregation, missing NaN handling,
or incomplete plots are considered more severe than stylistic issues.

The judge selects the candidate with the fewest and least severe defects,
preferring the simpler one when roughly equivalent. It returns only JSON with
three fields: \texttt{winner} (\texttt{"A"} or \texttt{"B"}), \texttt{reason}
(the decisive defect), and \texttt{loser\_critique} (one or two concrete
sentences on what the loser should fix next).

\endgroup

\noindent\textbf{D. Guardrail.}
\begingroup
\scriptsize

The guardrail receives a natural-language question \texttt{\{ML\_question\}},
the executed analyses and their Python code \texttt{\{ml\_results\}}, and any
previous errors \texttt{\{error\_guardrail\}}. Its role is to decide whether at
least one analysis correctly answers the question and, if so, to select the
best one.

Before deciding, the guardrail critiques each analysis against the same
checklist as the pairwise judge (valid columns; missing values handled before
modelling or aggregation; the metric or aggregation semantically required;
\texttt{result} at the correct grain; all required plots; the full question
answered). Plot completeness is judged from the code: if the question requires
multiple plots, e.g.\ one per variable, missing plots make the analysis
incorrect even if the numeric result is plausible.

The guardrail selects the analysis with the fewest and least severe defects.
Semantic errors such as wrong columns, wrong aggregation, missing NaN handling
that affects the result, or incomplete plots make an analysis incorrect. If
multiple analyses are valid, the guardrail prefers the simpler and more direct
one.

On success it returns JSON with \texttt{Correct\_response}=\texttt{"true"},
\texttt{Correct\_analysis\_number} (the selected index), and
\texttt{Detailed\_analysis}: a user-facing final report with the relevant
numbers and patterns, never mentioning code, rejected analyses, or the
selection process. On failure it returns
\texttt{Correct\_response}=\texttt{"false"}, \texttt{Advice} (the dominant root
cause), and \texttt{Per\_agent\_advice}: one concise, analysis-specific
critique each, citing real column names, avoiding code, and stating the
concrete fix for the next attempt.

\endgroup
\end{multicols}
}

\end{document}